\documentclass[a4paper]{cas-dc}
\ifdefined\pdfglyphtounicode
\input{glyphtounicode}
\fi
\usepackage[numbers,sort&compress]{natbib}
\usepackage{amsmath,amssymb,booktabs,array,tabularx,graphicx,microtype}
\usepackage{caption,hyperref,placeins,enumitem,xcolor}
\graphicspath{{figures/coverage20261027/}{figures/current/}}
\newcolumntype{Y}{>{\raggedright\arraybackslash}X}

\newcommand{\suite}{\mathcal{Q}_{41}}
\newcommand{\figref}[1]{Fig.~\ref{#1}}
\newcommand{\Framework}{M$^3$-Test}
\setlist[itemize]{leftmargin=1.2em,itemsep=2pt,topsep=3pt}

\begin{document}
\let\WriteBookmarks\relax
\shorttitle{M$^3$-Test: Coverage Gaps in Joint Robustness Testing}
\shortauthors{Wan and Wang}
\title[mode=title]{\texorpdfstring{M$^3$-Test}{M3-Test}: Coverage Gaps in Joint-Perturbation Testing of Visual Question Answering}
\author[1]{Enhui Wan}[orcid=0009-0000-1395-3212]
\ead{enhuiwan@bupt.edu.cn}
\cormark[1]\fnmark[1]
\credit{Conceptualization, Methodology, Software, Writing - original draft.}
\author[1]{Xiaowan Wang}[orcid=0009-0001-3779-7776]
\ead{wxw@bupt.edu.cn}
\fnmark[1]
\credit{Investigation, Data curation, Visualization, Writing - review \& editing.}
\affiliation[1]{organization={Beijing University of Posts and Telecommunications},addressline={Shahe Campus, No. 1 Nanfeng Road, Shahe Higher Education Park, Changping District},city={Beijing},postcode={102206},country={China}}
\cortext[1]{Corresponding author.}
\fntext[1]{These authors contributed equally to this work.}
\begin{abstract}
A visual question answering (VQA) model can pass every image-only and question-only test yet fail when the same changes occur together. Standard robustness reports average over conditions and therefore cannot say which accepted requests remain vulnerable. We introduce M$^3$-Test, a coverage audit that aligns each clean input with its single-modality variants and a complete $5\times5$ family, $4\times4$ strength joint library, so that any reduced test suite can be replayed on the same requests. Across 2.6 million responses from four vision--language models, joint failures affect 26--29\% of the Qwen questions that pass all 40 single-modality tests, although fewer than 0.6\% of their joint records fail. Passing the 25 maximum-strength corners does not close this gap: 11--14\% of the accepted questions still fail elsewhere in the library. Re-executing the archived failures reproduces about 90\% of them, against a 0.3--0.6\% flip rate among matched passing controls, and re-checking the most fragile failure records supports 80--89\% of them. On the evidence examined, neither execution noise nor grading error can fully account for the gap. The same aligned grids make previous evaluations reusable. On held-out images, a risk order learned from other models discovers about 80\% of affected Qwen questions within 25 additional queries, compared with 65--68\% for maximum-first and 53--57\% for pairwise covering arrays. M$^3$-Test thus turns a passing robustness suite into a request-level account of what remains untested, supporting acceptance, regression review and test allocation.
\end{abstract}
\maketitle

\section{Introduction}
A visual assistant may receive a blurred photograph together with a mistyped question. Before deploying the assistant, an engineer can test blurred images, test mistyped questions and inspect several severe combinations. If these tests pass, what has been established about the requests the system will receive? The answer depends on which combinations were inspected and which requests can fail outside that selection. An acceptance report therefore needs to connect a passing result to the behavior left untested.

Visual question answering (VQA) makes this problem concrete because its output depends on an image and a natural-language question~\cite{Antol2015VQA,Goyal2017VQAv2,Marino2019OKVQA,Hudson2019GQA}. Robustness benchmarks evaluate input degradations by family and strength~\cite{Hendrycks2019ImageNetC,Qiu2024MMRobustness,ishmam2025visual}. VQA consistency studies follow responses across related inputs~\cite{Jimenez2022CARETS,Chou2025MMR3,Rosenfeld2025VQAStability}. Recent testing approaches generate joint failures with matched single-modality controls or search image and text perturbation strengths~\cite{Duan2026VLTest,Weissl2026MoST}. These studies establish that both channels must be evaluated. None of them, however, answers the question an acceptance decision raises: once a test suite passes, which requests remain vulnerable to the combinations it did not run?

Aggregate scores cannot answer this question for three reasons. First, a joint error may repeat a failure already exposed by one channel, or appear only after both corresponding single inputs succeed; separating the two requires identical image and question realizations across all three tests. Second, a low failed-record rate can hide vulnerability spread across many requests, each failing under a few combinations. Third, the two channels vary in strength independently, so maximum-strength and equal-strength tests cover only specific coordinates of the joint space.

A complete library resolves all three. For each request, an acceptance suite can be replayed, its passes identified, and every accepted request checked against the rest of the library. This yields two complementary endpoints: discovery, the fraction of affected requests a suite finds, and residual failure, the fraction of accepted requests that still fail elsewhere. Both stay attached to concrete inputs, so the responsible combinations can be inspected and kept as regression tests.

\begin{figure*}[pos=tbp]
\centering\includegraphics[width=\textwidth]{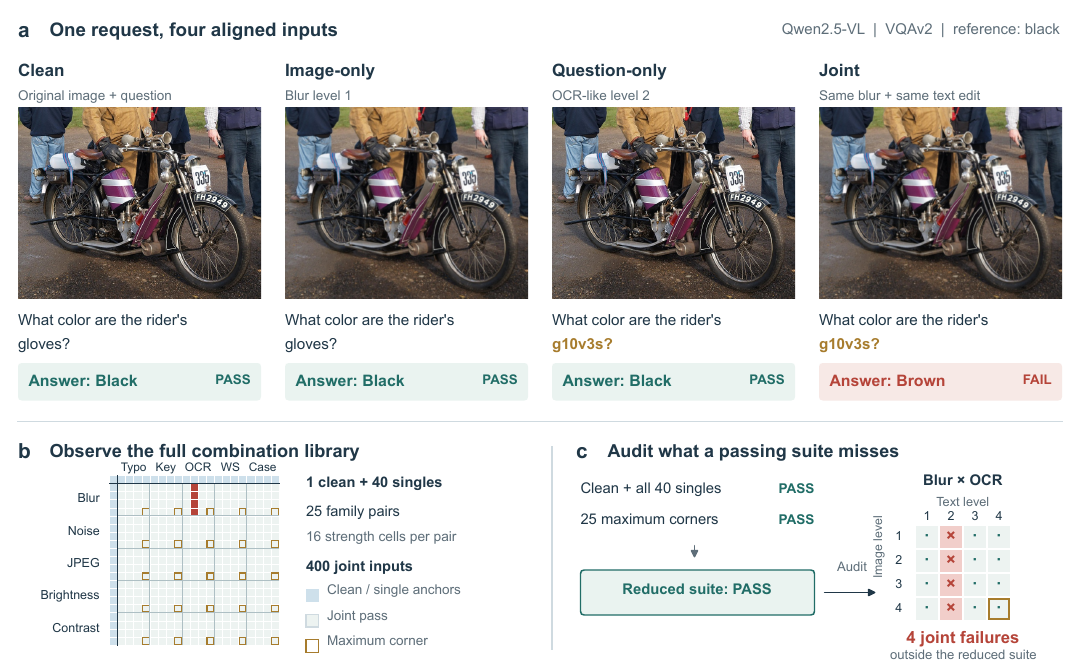}
\caption{A real request illustrates the M$^3$-Test coverage audit. (a) Original images, question strings and verbatim answers from Qwen2.5-VL for a VQAv2 question. Blur level 1 and OCR-like level 2 each preserve the correct answer; their combination changes it from Black to Brown. (b) All 441 observed response outcomes for the same request: blue cells are clean/single anchors, pale cells are joint passes, red cells are failures, and ochre outlines mark the 25 maximum-strength corners. Each family pair contains a $4\times4$ strength grid; WS denotes whitespace/punctuation. (c) Clean, all 40 singles and all 25 corners pass, yet four blur--OCR combinations fail. Following the same request across the full library exposes failures left outside the passing suite.}
\label{fig:framework}
\end{figure*}

We introduce \Framework{} --- the three Ms standing for multimodal inputs, multiple perturbation families and multiple strength levels --- a coverage audit built on this idea (\figref{fig:framework}). It aligns each clean input with its image-only and text-only variants and their Cartesian combinations: five families per channel and four strength levels give 400 joint conditions per image--question pair. Fully observed response grids allow single-modality screening, maximum-strength corners, diagonals and budgeted query orders to be compared on the same requests, and allow completed evaluations to guide where the next queries should go.

The study makes three contributions:
\begin{itemize}
\item \textbf{A request-level coverage audit for multimodal robustness.} Aligned clean, single and joint inputs with original-image identities and actual-input equivalence turn any reduced test suite into explicit discovery and residual-failure measurements with stated query costs.
\item \textbf{Evidence that passing tests leave large coverage gaps.} Across four models and 2.6 million responses, 21--29\% of the requests passing all single tests in LLaVA and both Qwen models fail under some joint condition, and 11--14\% of the Qwen requests passing all maximum-strength corners still fail. These findings hold under actual-input and semantic-only controls.
\item \textbf{Reusable risk orders for test allocation.} Coordinate risk learned from other models transfers to held-out images and raises 25-query discovery by 12.6--14.8 points over maximum-first for the Qwen models, which in turn outperforms pairwise covering arrays and uniform sampling.
\end{itemize}
Together, these results let an engineer decide, before accepting or replacing a model, which combinations threaten its established successes and how much a given test budget will reveal.

\section{Related Work}
\subsection{Corruption robustness and consistency}
ImageNet-C organizes common corruptions by family and severity; ImageNet-P follows prediction changes along perturbation sequences~\cite{Hendrycks2019ImageNetC}. Multimodal evaluations extend corruption analysis to image--text tasks, including clean-normalized degradation and adaptation strategies~\cite{Qiu2024MMRobustness,Chen2023AdaRobustness,Latif2026VLSRB}. Visual VQA robustness also incorporates error range and severity slope~\cite{ishmam2025visual}, while Shirnin et al. analyze responses to noise in both inputs~\cite{Shirnin2024Robustness}. Text-side work supplies the corresponding vocabulary: synthetic character noise breaks neural machine translation~\cite{Belinkov2018SyntheticNoise}; EDA studies word-level substitutions, insertions, swaps and deletions~\cite{Wei2019EDA}, while AEDA inserts punctuation~\cite{Karimi2021AEDA}. Adversarially edited questions probe a human-driven difficulty axis~\cite{Li2021AdversarialVQA}, and debiasing methods improve robustness to language priors~\cite{Cao2024LanguagePriors,Peng2024PolarityVQA}, within a broader robust-VQA literature of datasets and mitigation methods~\cite{ma2024robust,Chen2020CSS,Kant2021ContrastClassify} and adversarial-robustness designs for vision--language models~\cite{Bhagwatkar2024VLMRobustness}. These works characterize average, extreme or severity-dependent performance. Robustness is also read through calibration and reliability, since accuracy can hide miscalibration~\cite{Guo2017Calibration} and selective-prediction studies decide which VQA responses to trust~\cite{Dancette2023SelectiveVQA,Khan2024SelectiveVQA}, while character-level adversarial methods search for flipping edits directly~\cite{Ebrahimi2018HotFlip} and evaluation benchmarks probe related distribution shifts in VQA and in omni-modal reasoning~\cite{Sterz2025DARE,Kang2026SCEval}. Scene-text VQA evaluates reading over images that themselves contain text~\cite{Singh2019TextVQA,Biten2019STVQA}, with text-based reasoning modelling question--scene-text relations~\cite{Zhou2023SparseTextVQA,Zhou2024GPIN} and extending to egocentric video~\cite{Zhou2025EgoTextVQA,Zhou2026GroundedTextVQA}; such datasets, and VizWiz~\cite{Gurari2018VizWiz}, are excluded here so that question-string edits are not confounded with recognising text inside the image. Our audit instead follows the individual requests accepted by a reduced suite and measures the failures they still carry in the full two-axis library.

Consistency evaluation supplies an important foundation for retaining input identity. VQA-Rephrasings studies equivalent questions~\cite{shah2019cycle}; CARETS uses paired capability tests and comprehensive accuracy~\cite{Jimenez2022CARETS}; MM-R$^3$ evaluates accuracy and consistency under rephrasing, restyling and context reasoning~\cite{Chou2025MMR3}. Stability analysis further relates benign transformations to answer correctness~\cite{Rosenfeld2025VQAStability}. These works show why aligned variants matter. M$^3$-Test uses that alignment to check each joint coordinate against its corresponding successful singles and audit coverage across the resulting grid.

\subsection{Joint failure generation and modality sensitivity}
The preprint VLTest generates visual and textual variants, checks answer preservation and attributes joint failures through four matched inputs~\cite{Duan2026VLTest}. Its output-score analysis also distinguishes additive from interaction failures. The preprint MoST searches continuous perturbation strengths for object localization, reporting modality sensitivity and degradation relative to perturbation effort~\cite{Weissl2026MoST}. Both treat joint inputs, adjustable strengths and modality-specific analysis as explicit testing objects.

VLTest also reports SeedCov, the fraction of admitted seeds exposing at least one confirmed failure, and charges deduplicated input keys~\cite{Duan2026VLTest}. M$^3$-Test moves from failure generation to acceptance auditing. A fully observed family--strength library serves as a common reference against which any reduced suite can be replayed. Discovery counts the affected questions a suite finds, and residual failure counts those it accepts but should not, making omissions after maximum-strength or diagonal passes directly measurable. The same input identities also support held-out-image evaluation of historical query orders.

\subsection{Combinatorial coverage and test adequacy}
Behavioral testing organizes model evaluation around capabilities and targeted test types, as illustrated by CheckList~\cite{Ribeiro2020CheckList}. Combinatorial testing uses covering arrays to exercise every value combination of selected $t$-factor subsets~\cite{Kuhn2010Combinatorial}. The factorization determines the guarantee: treating the image variant and text variant as two factors requires their full Cartesian product for two-way coverage; treating family and level as separate factors permits smaller arrays covering lower-order interactions. Our complete library supplies an observed reference for measuring which affected requests a reduced suite discovers and which failures remain among its passes.

DeepCT studies combinatorial interactions among neural activations, while DeepGauge defines multi-granularity activation-coverage criteria~\cite{Ma2019DeepCT,Ma2018DeepGauge}. Subsequent evaluation shows that increasing neuron coverage can diverge from effective defect detection~\cite{HarelCanada2020Coverage}. We therefore evaluate a suite by the response failures it exposes on requests whose reference inputs succeed, rather than by input coverage alone.

\subsection{Test prioritization and preservation}
Neural-network test prioritization allocates limited evaluation effort to promising inputs. DeepGini uses output impurity~\cite{Feng2020DeepGini}; replication work highlights the importance of simple, competitive baselines~\cite{Weiss2022Simple}. We prioritize at the level of perturbation coordinates instead, using failure-frequency and greedy-coverage orders learned from completed response grids on disjoint images, including grids of other models.

Model-update studies identify negative flips, where a previously correct prediction becomes incorrect~\cite{Yan2021PositiveCongruent}. Tracking success over a whole input suite rather than a single input extends this view to joint robustness, and Appendix~\ref{app:changes} applies it to a model replacement.

\section{Coverage Audit and Study Design}
\subsection{A paired finite input library}\label{sec:library}
Each question $k$ contains image $I_k$, question $Q_k$, acceptable references $\mathcal A_k$ and original-image identity $g_k$. For image family $a$, text family $b$ and levels $i,j$, a joint input is
\begin{equation}
x_k^{a,b}(i,j)=\bigl(T^I_{a,i}(I_k),T^Q_{b,j}(Q_k)\bigr).
\end{equation}
Level zero preserves the channel. The corresponding singles are $x_k^{a,b}(i,0)$ and $x_k^{a,b}(0,j)$, reusing the same image bytes and question strings as the joint input. Each question has one clean input, 20 image-only conditions, 20 text-only conditions and $25\times16=400$ joints. We use $c$ to index these joint coordinates and write $y^I_{mkc}$, $y^T_{mkc}$ and $y^J_{mkc}$ for the binary outcomes of its image-only anchor, text-only anchor and joint input under model $m$.

\begin{figure*}[pos=tbp]
\centering\includegraphics[width=\textwidth]{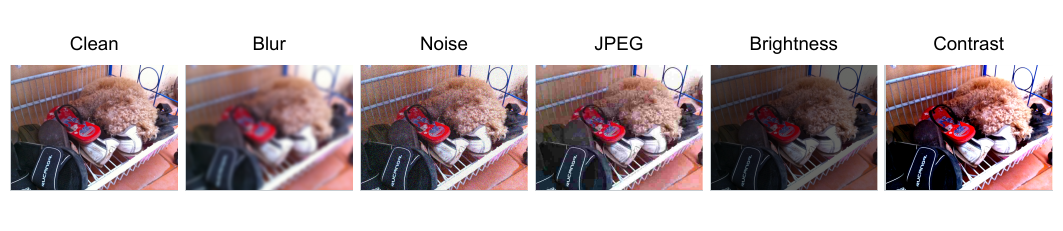}
\caption{Clean and maximum-level image illustrations for the five operators. Corresponding question examples appear immediately below. Parameters and calibration measurements are given in Appendix~\ref{app:operators}.}
\label{fig:inputs}
\medskip\captionsetup{hypcap=false}
\captionof{table}{Maximum-level question variants. Red characters mark edits; visible spaces expose repeated spaces.}
\label{tab:text_examples}\small\setlength{\tabcolsep}{4pt}
\begin{tabular}{@{}ll@{}}\toprule
Family & Question\\\midrule
Clean & How long has ultimate frisbee been an official sport?\\
Typo & How l\textcolor{red}{no}g has ultimate fr\textcolor{red}{r}isbee been an offic\textcolor{red}{ai}l sp\textcolor{red}{O}rt?\\
Keyboard & How long has ultimate fri\textcolor{red}{d}bee be\textcolor{red}{w}n an o\textcolor{red}{v}ficial sport?\\
OCR-like & How long has u\textcolor{red}{1}timat\textcolor{red}{3} frisbee b\textcolor{red}{33}n an off\textcolor{red}{1}c\textcolor{red}{1}al sport?\\
Whitespace/punc. & How\textvisiblespace\textcolor{red}{\textvisiblespace}long\textvisiblespace\textcolor{red}{\textvisiblespace}has\textvisiblespace\textcolor{red}{\textvisiblespace}ultimate\textvisiblespace\textcolor{red}{\textvisiblespace}frisbee\textvisiblespace\textcolor{red}{\textvisiblespace}been\textvisiblespace\textcolor{red}{\textvisiblespace}an official sport\\
Casing & How long \textcolor{red}{H}as ulti\textcolor{red}{M}ate frisbee been an \textcolor{red}{O}fficial sport?\\\bottomrule
\end{tabular}
\end{figure*}

Image families are Gaussian blur, Gaussian noise, JPEG compression, brightness and contrast. Text families are typographical edits, keyboard-neighbor substitutions, OCR-like character confusion, whitespace/punctuation changes and casing. The image families reflect capture and transmission degradations; the text families reflect questions entered on mobile keyboards, transcribed from speech or extracted by OCR, where OCR-like edits apply character confusions to the question string. Meaning preservation is enforced by the operators rather than checked afterwards. Interrogative, quantity, direction and negation words, every token containing a digit, and every token shorter than four characters are never edited. Typographical and keyboard edits touch only interior characters, so each word keeps its first and last character and at most one internal position changes; OCR-like edits keep the first character and every digit; and the question mark survives below the maximum level. A stratified audit of 300 perturbed questions, five in each text family, level and source dataset, judged by two annotators, found meaning, answer intent and readability preserved in all 300 (Wilson 95\% lower bound 0.987), so the families change surface form without changing what is asked (Appendix~\ref{app:operators}). \figref{fig:inputs} and Table~\ref{tab:text_examples} show both channels together; Appendix~\ref{app:operators} gives parameters and input-magnitude calibration.

The 1,500 questions are sampled equally from processed GQA, OK-VQA and VQAv2 validation pools~\cite{Hudson2019GQA,Marino2019OKVQA,Goyal2017VQAv2}. Seed-zero selection prioritizes image fingerprints and balances question types within source quotas. They represent 1,065 original-image groups. Asset realizations are determined by parent identity, family, level and seed. Levels order control magnitude within each family. Each level is realized independently: higher-level text edits can fall at different positions, and brightness/contrast directions can differ between levels. A maximum-level corner is therefore the highest-magnitude realization of each operator, not a superset of its lower-level edits.

\subsection{Models, judgments and input identity}\label{sec:judgments}
We analyze BLIP-2 OPT-2.7B, LLaVA-1.5-7B, Qwen2.5-VL-7B-Instruct and Qwen3-VL-8B-Instruct~\cite{Li2023BLIP2,Liu2024LLaVA15,Bai2025Qwen25VL,Bai2025Qwen3VL}. All receive the same frozen inputs and the instruction ``Answer the following question with a short, direct answer. Do not include any explanation.'' Responses are generated with model-specific processors and greedy decoding, using seed 0 and at most 32 new tokens. Their 2,646,000 free-text responses are evaluated under one binary protocol.

Response success requires semantic agreement with an original reference and compliance with the answer instruction:
\begin{equation}\label{eq:task}
z_{mkc}=s_{mkc}f_{mkc},\qquad s_{mkc},f_{mkc}\in\{0,1\}.
\end{equation}
A restricted safe-exact matcher handles exact-reference answers after normalizing case, whitespace, outer quotation marks and trailing sentence punctuation, while preserving word order, negation, signs and internal punctuation. Other responses receive independent semantic and format judgments from the same frozen Qwen3-4B-Instruct-2507 Judge. Semantic grading uses the original question, full references and prediction; format grading uses the instruction and prediction. Equivalent meanings and brief direct sentences are allowed. Accuracy is the fraction of records with $z_{mkc}=1$, reported over the 1,500 clean, 60,000 single and 600,000 joint nominal records of each model (Table~\ref{tab:summary}); each record contributes once even when it duplicates an equivalent input. All coverage endpoints below instead count each question once.

Because LLM-based evaluation can exhibit evaluator-dependent preferences~\cite{Zheng2023LLMJudge,Panickssery2024SelfPreference}, every model is graded with the same reference-based binary questions, and semantic and format outcomes are kept separate so that their contributions to each endpoint can be isolated (Appendix~\ref{app:controls}).

Perturbations can leave an input unchanged, for example when a probabilistic edit selects no token in a short question. An input-equivalence audit therefore compares decoded RGB pixels and exact question strings within each question: each model's 661,500 nominal entries correspond to 384,069 distinct actual inputs, and equivalent entries share one identity for coverage and query accounting. As a control, we additionally require both channels to change and exclude the small set of equivalent groups whose responses carry conflicting success labels (Appendix~\ref{app:controls}). The initial cohort stays fixed throughout.

Two further checks address the failure evidence itself. In an answer review, the authors re-examined a targeted set of archived joint failures together with matched passing controls. In an execution repeatability study, every archived joint failure of the two Qwen models and an equal number of matched passing controls were re-executed once on the original environment, with one request per batch. Appendix~\ref{app:repeat} reports both.

\subsection{Screening and coverage endpoints}\label{sec:endpoints}
Let $y^C_{mk}$ be clean success. Corresponding-single eligibility at coordinate $c$ is
\begin{equation}\label{eq:eligible}
E_{mkc}=y^C_{mk}\,y^I_{mkc}\,y^T_{mkc}.
\end{equation}
The clean anchor ensures that a conditional joint failure concerns a request the model already handled before either change. Family-pair maps pool $E_{mkc}(1-y^J_{mkc})$ and $E_{mkc}$ across the pair's 16 coordinates; their ratio is the pair's conditional joint failure rate.

For acceptance auditing, define the set passing clean and all single conditions
\begin{equation}
\suite^m=\{k:y^C_{mk}=1\ \land\ \text{all 40 single outcomes pass}\}.
\end{equation}
Within this fixed set, $F_{mk}=1$ if any of the 400 joint conditions fails. The event density averages failed joint records, while the affected-question fraction averages $F_{mk}$. Each request contributes once to the latter, irrespective of how many conditions fail.

For a selected joint suite $\mathcal T$, let $G_{mk}(\mathcal T)=1$ when every selected test passes. Discovery and residual failure are
\begin{align}
D_m(\mathcal T)&=\frac{\sum_{k\in\suite^m}F_{mk}[1-G_{mk}(\mathcal T)]}{\sum_{k\in\suite^m}F_{mk}},\label{eq:discovery}\\
R_m(\mathcal T)&=\frac{\sum_{k\in\suite^m}F_{mk}G_{mk}(\mathcal T)}{\sum_{k\in\suite^m}G_{mk}(\mathcal T)}.\label{eq:residual}
\end{align}
$D_m$ measures the affected questions found; $R_m$ measures the fraction of accepted questions failing elsewhere in the library, and thus characterizes the model, library and suite jointly. Maximum-corner testing uses $(4,4)$ in each family pair (25 joints); diagonal testing uses all $(i,i)$ coordinates (100 joints).

\subsection{A model fragility profile}\label{sec:profile_method}
To compare models, we report three conditional accuracies: image-only success after clean success, text-only success after clean success, and joint success after the corresponding singles succeed.

For matched comparison, $\mathcal C$ contains the 469 questions all four models answer correctly on clean input. Image and text retention each use their 20 variants over this same set. Joint comparison fixes the 141,396 records whose clean and corresponding single inputs all four models pass. Writing $E^*_{kc}=\prod_m E_{mkc}$, joint retention is
\begin{equation}\label{eq:retention}
P^J_m=\frac{\sum_{k,c}E^*_{kc}y^J_{mkc}}{\sum_{k,c}E^*_{kc}}.
\end{equation}
The text-minus-image failure difference equals image retention minus text retention; its sign identifies the weaker channel.

Marginal scores do not determine paired risk. Consider four clean-successful requests with image outcomes $(1,1,0,0)$ and text outcomes $(1,0,1,0)$ at one coordinate. Joint vectors $(1,0,1,0)$ and $(0,1,1,0)$ have identical mean accuracy, but respectively preserve and break the only request on which both singles succeed. Likewise, equal per-cell accuracies can concentrate errors on one question or spread them over several, changing whole-suite survival. Aligned outcomes retain exactly this information.

\subsection{Query budgets and historical risk}\label{sec:orders}
Budgeted replay uses the fixed population passing clean and all singles under the combined success criterion. We compare maximum-first, diagonal-first, exact uniform-without-replacement expectation, historical failure frequency and greedy coverage. The maximum order prioritizes the sum of levels, then their minimum and image level, with lexical ties; diagonal-first prioritizes equal levels before the same ordering. Frequency ranks coordinates by the calibration set's failure counts. Greedy coverage selects coordinates exposing the most not-yet-covered calibration questions for its first 100 positions, followed by frequency order. Ties use maximum-first.

A combinatorial baseline treats image family, image level, text family and text level as four factors. Each pairwise covering array contains 25 coordinates and covers every value combination of every two-factor projection. We construct 64 outcome-independent arrays from a five-symbol orthogonal array, mapping the two level factors onto four levels with balanced repeated-level choices, seed 17. Each order tests its array coordinates in maximum-first order, then follows maximum-first over remaining coordinates. Twenty-five rows is the minimum for strength-two coverage here, since the two five-valued family factors alone have 25 value pairs. Discovery is averaged per request across all arrays before image resampling, and variation across designs is reported separately. These orders require no response calibration.

Original-image groups are partitioned into five folds with seeds 17, 29 and 43. Every model's questions and conditions for a group follow the same fold. Each target fold is evaluated with orders fitted on the other four folds. Source policies use either the target model's calibration results, one of the other three models, or their predeclared pooled failure counts. The pooled-other policy uses no target-model calibration joint responses. Leave-one-source-out replay additionally fits on two datasets and evaluates on the third.

Additional joint budgets are 10, 25, 50, 100 and 200 queries per question passing the initial tests. We retain nominal-coordinate accounting and distinct-actual-input accounting separately. Distinct-input replay skips inputs equivalent to measured clean/single anchors or previous joint inputs using identity alone, then continues through the order until the budget is filled or the unknown inputs are exhausted. Maximum-first at 25 distinct queries therefore draws from the continuing order, whereas the nominal corner gate fixes the 25 $(4,4)$ coordinates. Every queried input is charged, and controlled-failure discovery uses label-consistent groups.

For uniform sampling, if a question has $n$ available inputs and $h$ failure inputs, its discovery probability at budget $B$ is
\begin{equation}\label{eq:uniform}
1-\frac{\binom{n-h}{\min(B,n)}}{\binom{n}{\min(B,n)}}.
\end{equation}
Summing these probabilities gives expected discovery.

\subsection{Uncertainty and cost accounting}\label{sec:uncertainty}
We use 2,000 shared bootstrap draws of the 1,065 original-image groups, seed 17. Each sampled group carries all its questions, conditions and models. Ratios are recomputed from the resampled numerator and denominator; pointwise 95\% intervals use linear percentile quantiles. Paired differences use the same draws. For the four model-specific history--maximum comparisons at budget 25 in Section~\ref{sec:result_transfer}, we additionally report Bonferroni-adjusted 98.75\% intervals as a four-comparison sensitivity analysis. Other intervals are pointwise, and the initial cohorts remain fixed. For budget replay, each question's held-out predictions are averaged over the three partition seeds before image resampling, so the intervals describe variation across images given the fitted orders.

Cost has three parts: calibration, target screening and additional joint queries. Screening charges every evaluated root, including those rejected by qualification. Calibration includes the source models' screening and the complete joint records required to fit the order. Distinct-input accounting applies to both stages. We report incremental deployment calls and calibration-plus-deployment calls by model, measuring acquisition and screening in distinct VLM input queries.

Library-size sensitivity holds the original population passing clean and all singles fixed. For each choice of $r$ image families and $s$ text families, we exhaust all family subsets and retain their complete strength grids, giving $16rs$ joint coordinates per sublibrary. The analysis covers 961 sublibraries. Mean affected fractions and original-image intervals summarize these selections; the minimum and maximum across equal-size family choices separately describe library composition.

Failure multiplicity counts the failed joint coordinates $N_{mk}$ of each question on the same fixed cohort. We report the fraction with $N_{mk}\geq2$ and repeat the count over distinct changed inputs with consistent success labels. As a reference, a shared-rate independent model draws $N\sim\mathrm{Binomial}(400,p)$ with $p=1-(1-n_A/n_Q)^{1/400}$, fitted to the observed affected count $n_A$ in a cohort of $n_Q$ questions. It matches the zero-failure fraction by construction, so the comparison isolates how failures are distributed among affected questions.

\section{Results}
\subsection{Sparse joint failures affect many requests passing single tests}\label{sec:result_affected}
Passing all single conditions left substantial joint vulnerability. Among the Qwen models' questions passing clean and all singles, more than one quarter failed under at least one combination (Table~\ref{tab:summary}), even though fewer than 0.6\% of their joint records failed. A favorable event average thus coexists with a large set of requests that do not survive the full library.

\begin{table*}[pos=tbp]
\centering\small\setlength{\tabcolsep}{3.5pt}
\caption{Response accuracy and joint vulnerability. Clean/single/joint accuracy uses 1,500/60,000/600,000 records per model. The final columns use each model's own fixed clean-plus-all-single cohort. Event density divides failed joint records by 400 times the qualified-question count; affected questions are counted once. Affected fractions include original-image bootstrap 95\% intervals.}
\label{tab:summary}
\begin{tabular}{@{}lrrrrrr@{}}\toprule
Model & Clean(\%) & Single(\%) & Joint(\%) & Passed($n$) & Event density(\%) & Affected(\%) [95\% interval]\\\midrule
BLIP-2 & 41.33 & 40.00 & 38.58 & 145 & 1.048 & 61.38 [53.24, 69.94]\\
LLaVA & 69.13 & 67.90 & 66.63 & 550 & 0.300 & 21.27 [17.84, 24.71]\\
Qwen2.5-VL & 71.47 & 69.38 & 67.28 & 540 & 0.514 & 28.70 [25.14, 32.61]\\
Qwen3-VL & 75.60 & 74.28 & 72.96 & 658 & 0.417 & 25.84 [22.51, 29.48]\\
\bottomrule\end{tabular}
\end{table*}

Most affected questions fail under more than one combination. For both Qwen models, about one fifth of the questions passing clean and all single tests fail at two or more joint coordinates, and this fraction moves by at most 0.3 percentage points when failures must come from distinct inputs with both channels changed and consistent labels (21.85\% to 21.85\% and 20.36\% to 20.06\%; Table~\ref{tab:multiplicity}). Compared with a shared-rate independent reference at the same affected fraction, multiple-failure questions are 4.8 and 5.6 times as common for Qwen2.5-VL and Qwen3-VL. Joint failures therefore concentrate on particular requests, which is precisely what request-level coverage captures and event density averages away.

\begin{table*}[pos=tbp]
\centering\small\setlength{\tabcolsep}{5pt}
\caption{Failure-evidence multiplicity on fixed cohorts passing clean and all 40 single conditions: 540 questions for Qwen2.5-VL and 658 for Qwen3-VL. Observed nominal counts use the 400 joint coordinates; the shared-rate reference is fitted to their affected fraction, so zero-failure agreement is by construction. The last row for each model counts distinct actual inputs after requiring both channels to change and success labels to be consistent. All entries are percentages of the same model-specific cohort.}
\label{tab:multiplicity}
\begin{tabular}{@{}llrrr@{}}\toprule
Model & Evidence / reference & No failure & One failure & At least two failures\\\midrule
Qwen2.5-VL & Observed coordinates & 71.30 & 6.85 & 21.85\\
Qwen2.5-VL & Shared-rate reference & 71.30 & 24.13 & 4.57\\
Qwen2.5-VL & Distinct changed inputs & 71.30 & 6.85 & 21.85\\
Qwen3-VL & Observed coordinates & 74.16 & 5.47 & 20.36\\
Qwen3-VL & Shared-rate reference & 74.16 & 22.18 & 3.66\\
Qwen3-VL & Distinct changed inputs & 74.16 & 5.78 & 20.06\\
\bottomrule\end{tabular}
\end{table*}

Which families are tested matters as much as how many. Balanced three-family subsets expose on average about three fifths of the Qwen requests affected by the full library, and equal-size subsets differ substantially with the chosen families (Appendix~\ref{app:library_size}).

The affected sets are robust to input identity and grading. Requiring both channels to change and excluding label-discordant equivalent inputs leaves the affected-question set of every model exactly unchanged (Appendix~\ref{app:controls}). Removing format compliance from scoring also preserves the coverage gap: under semantic-only requalification, residual failure after the maximum corners is 11.41--15.40\% for the two Qwen models (Table~\ref{tab:semantic_coverage}).

The failure evidence survives re-execution and manual review. Re-running every archived Qwen joint failure alongside an equal number of matched passing controls reproduced 89.83\% and 90.44\% of the failures, whereas only 0.27\% and 0.64\% of the matched controls flipped to failure (Appendix~\ref{app:repeat}). Nearly every affected question still failed under at least one condition (97.42\% and 94.71\%; 151/155 and 161/170), and 84.43\% and 85.61\% of the archived failures kept all three anchors correct (938/1,111 and 940/1,098), covering 148 and 157 of those questions. The answer review agrees: of 71 archived failures, 63 (88.7\%) were retained as failures under the original labels, and 57 (80.3\%) were supported as valid joint failures once synonym, reference and format disputes were resolved; 4 records (5.6\%) were answers the Judge should have accepted, and no matched control was confirmed as a hidden failure. Execution variability and grading error are therefore far too small to account for the coverage gap.

\begin{figure*}[pos=tbp]
\centering\includegraphics[width=\textwidth]{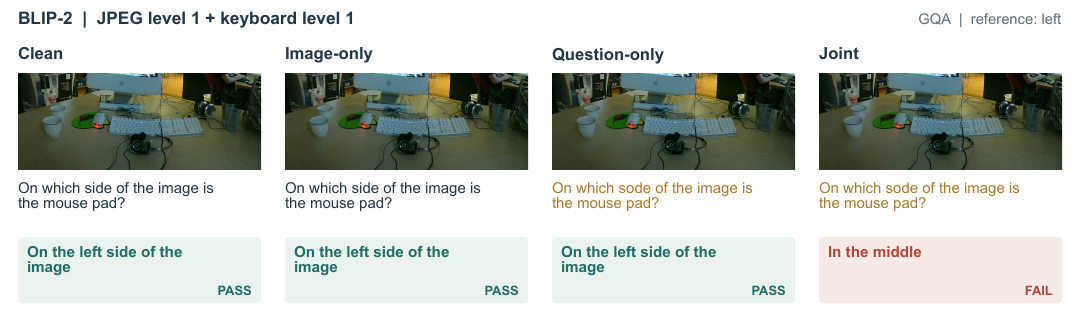}
\caption{A further observed four-input failure: BLIP-2 under JPEG level 1 and keyboard level 1, for a GQA question. The question edit changes ``side'' to ``sode.'' Clean and both singles pass, whereas the joint answer changes the mouse pad's location from left to middle. Images, question strings and answers are taken from the case records; this example and the case in Fig.~\ref{fig:framework} were selected from 3,237 eligible events.}
\label{fig:cases}
\end{figure*}

Individual failures are directly inspectable. In Fig.~\ref{fig:cases}, BLIP-2 answers the clean and both single inputs correctly, yet the joint input moves the mouse pad from left to middle; together with the Qwen2.5-VL case in Fig.~\ref{fig:framework}, it shows how each failure traces back to its exact perturbations.

\subsection{Reduced joint suites leave failures among their passes}\label{sec:result_gates}
\begin{figure*}[pos=tbp]
\centering\includegraphics[width=\textwidth]{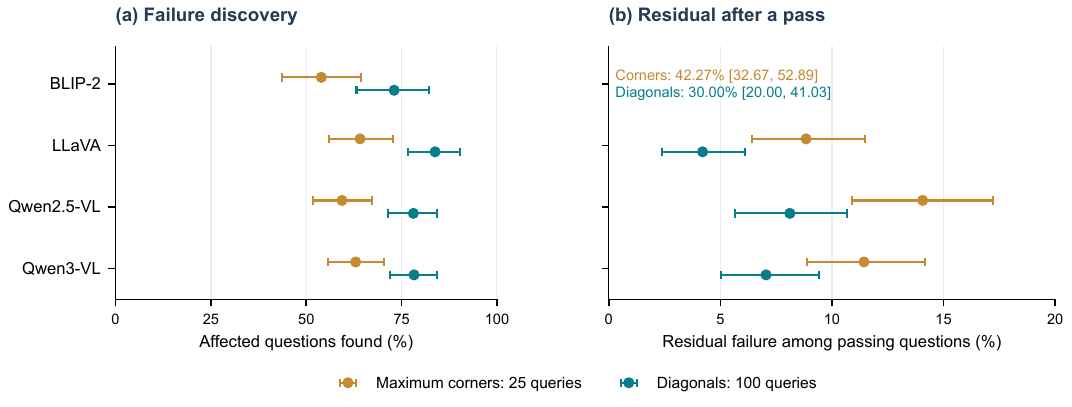}
\caption{Discovery and residual failure describe different consequences of a reduced test suite. (a) Fraction of full-library-affected questions discovered, Eq.~\eqref{eq:discovery}. (b) Fraction of passing questions that fail elsewhere, Eq.~\eqref{eq:residual}. The plotted residual scale resolves LLaVA and the Qwen models; BLIP-2 estimates lie above that scale and are given numerically on its row with complete 95\% intervals. Each model uses its own cohort passing clean and all singles. Error bars are pointwise original-image bootstrap 95\% intervals. Maximum corners and diagonals cost 25 and 100 nominal joints per qualified question.}
\label{fig:gates}
\end{figure*}

Maximum-strength testing exposed many affected questions but left measurable risk among the requests it accepted. After clean, all 40 singles and all 25 maximum corners passed, residual failure was 14.06\% for Qwen2.5-VL and 11.43\% for Qwen3-VL (\figref{fig:gates}). Even with four times the budget, complete diagonal testing still left 8.11\% and 7.05\% (exact counts in Appendix~\ref{app:gate_counts}).

The discovery view shows where this residual risk comes from. Maximum corners missed about two fifths of affected Qwen2.5 questions and over one third of affected Qwen3 questions; diagonals missed roughly one fifth in both models. Many missed failures lie just off the corner: the coordinate $(4,3)$ alone exposed more than half of Qwen2.5's corner-missed questions. Off-corner realizations therefore add substantial request-level coverage.

\begin{figure*}[pos=tbp]
\centering\includegraphics[width=\textwidth]{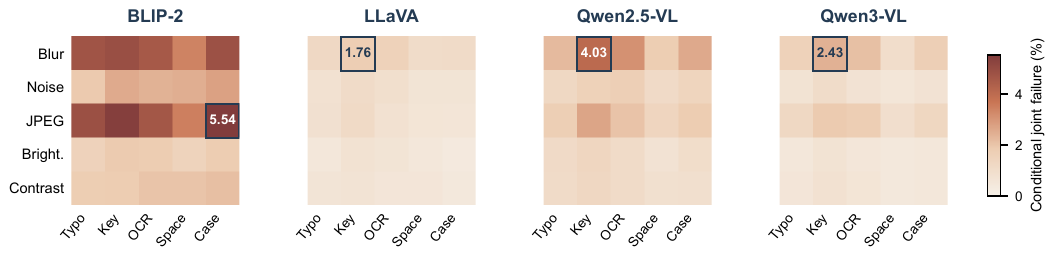}
\caption{Clean-anchored joint risk after both corresponding single inputs succeed. Each family-pair cell pools its 16 strength coordinates; color encodes failed/eligible records on a common scale. Space denotes whitespace/punctuation (WS in Fig.~\ref{fig:framework}). The outlined cell is each model's maximum. Each model uses its own clean-and-single-qualified population; the map localizes high-risk combinations within that population.}
\label{fig:pair_risk}
\end{figure*}

The condition map shows where to look (\figref{fig:pair_risk}). Blur combined with keyboard substitutions had the highest conditional failure rate in both Qwen models, at 4.03\% and 2.43\%. Such hotspots are natural priorities for regression suites and for perturbation-aware data augmentation, and each traces back to the requests involved.

\subsection{Matched profiles distinguish failure dimensions}\label{sec:result_profiles}
\begin{figure*}[pos=tbp]
\centering\includegraphics[width=\textwidth]{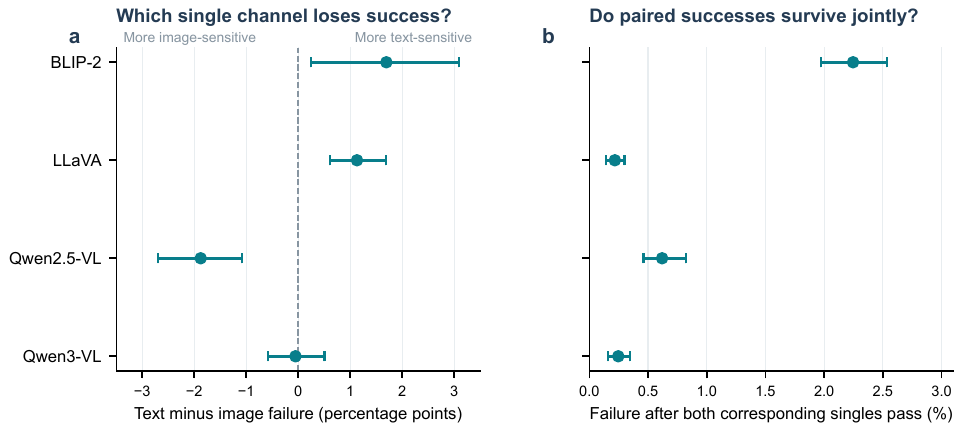}
\caption{Failure dimensions under matched qualification. (a) Text-minus-image failure on 469 common clean-successful questions, with 20 variants per channel. Positive values indicate more text failures in this library. (b) Joint failure on the same 141,396 exact-coordinate records whose anchors all four models pass. Points and 95\% intervals use original-image resampling.}
\label{fig:profile}
\vspace{3pt}
\centering\includegraphics[width=\textwidth]{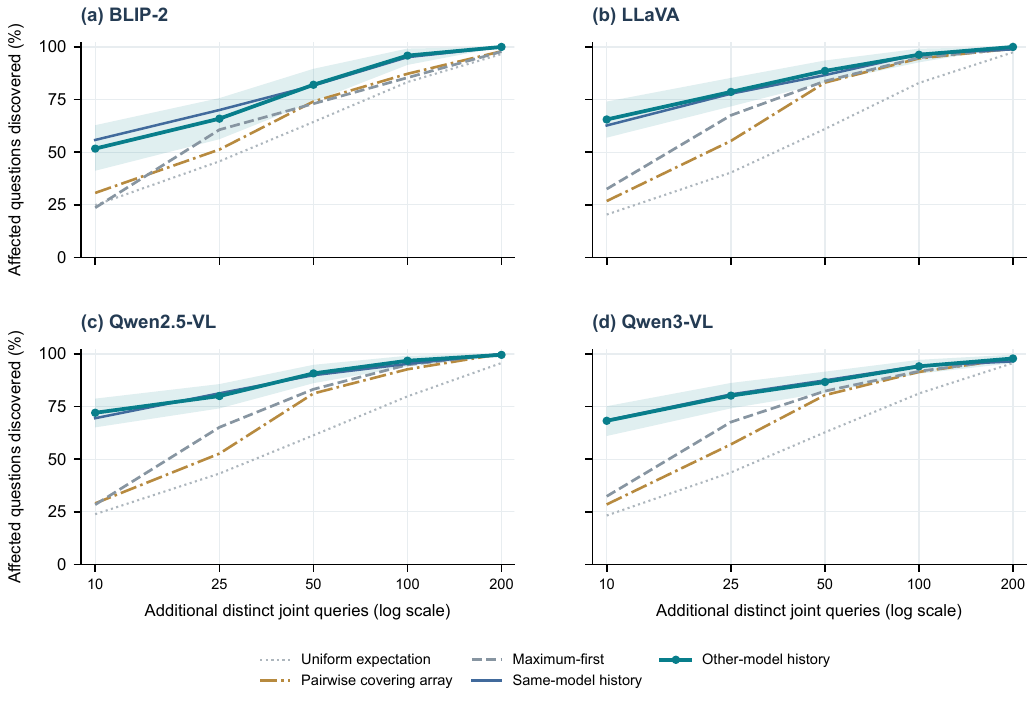}
\caption{Discovery as the distinct-input joint budget grows. Target and source-model outcomes from held-out images are excluded from fitting all risk orders. The pairwise covering-array curve averages 64 outcome-independent designs. Shading gives pointwise original-image 95\% intervals for pooled-other-model history; uniform is the exact without-replacement expectation. The horizontal axis is logarithmic. Initial screening and pilot acquisition costs are additional to the plotted joint budget.}
\label{fig:transfer}
\end{figure*}

On the same clean-successful questions, models showed different weak channels. LLaVA's text failure rate exceeded its image failure rate by 1.13 percentage points, with a paired 95\% interval of $[0.61,1.68]$. Qwen2.5-VL showed the opposite pattern: image failures exceeded text failures by 1.88 points, interval $[1.09,2.70]$ (\figref{fig:profile}). Both directions held under semantic-only scoring, while Qwen3-VL was balanced between channels.

Conditional joint retention can also reverse the ranking implied by raw accuracy. Qwen2.5-VL's unconditional joint accuracy exceeded LLaVA's, yet on the matched records whose anchors all models pass it retained 0.402 percentage points fewer, paired interval $[0.215,0.607]$; Qwen3-VL likewise retained more than Qwen2.5-VL. Broad task capability and preservation of demonstrated success are thus distinct properties, and the channel contrast indicates where augmentation is most likely to help, for example visual degradations for Qwen2.5-VL. Appendix~\ref{app:profiles} gives all retention scores and semantic views.

\subsection{Historical risk improves equal-budget discovery}\label{sec:result_transfer}
Historical response grids transfer to held-out images and to other target models. At 25 additional distinct joint queries, the order learned from the other three models discovered about 80\% of affected questions for both Qwen models, compared with roughly two thirds for maximum-first (Table~\ref{tab:transfer}). The paired improvements were 14.84 and 12.55 percentage points, and both stay above zero under Bonferroni-adjusted 98.75\% intervals ([5.95, 24.61] and [5.06, 20.77]). LLaVA improved by 11.11 points (adjusted interval [0.14, 22.64]); for BLIP-2, whose small cohort widens all intervals, the gain of 5.24 points was not resolved ([-6.67, 18.24]). Other-model history matched same-model history within 1.3 points for LLaVA and both Qwen models, so a new model can be screened efficiently without any of its own calibration responses.

Among calibration-free orders, maximum-first was the strongest. It exceeded the mean pairwise covering-array design for every model, by 10.60--12.51 points for the Qwen models, and lay 15.0--27.2 points above the exact uniform expectation. Covering every pair of input factors and discovering response failures are distinct objectives: observed failures occupy particular four-factor coordinates, which strength ordering reaches sooner.

\begin{table*}[pos=tbp]
\centering\fontsize{8.5}{10}\selectfont\setlength{\tabcolsep}{3pt}
\caption{Affected-question discovery at 25 distinct joint queries. Failure discovery uses distinct inputs with both channels changed and consistent labels. Uniform is the exact without-replacement expectation; its brackets give pointwise image-bootstrap 95\% intervals. Covering-array results average 64 outcome-independent designs. Same-model and pooled-other-model history use only calibration images. Differences compare other-model history with maximum-first, using pointwise paired image-bootstrap 95\% intervals. Affected-question denominators are 89/117/155/170; historical predictions average three partition seeds per question.}
\label{tab:transfer}
\begin{tabular}{@{}lrrrrrr@{}}\toprule
Target & \shortstack{Uniform(\%)\\{[95\% interval]}} & \shortstack{Maximum-\\first(\%)} & \shortstack{Covering\\array(\%)} & \shortstack{Same-model\\history(\%)} & \shortstack{Other-model\\history(\%)} & \shortstack{History--maximum difference\\{[95\% interval], pp}}\\\midrule
BLIP-2 & 45.65 [39.77, 51.60] & 60.67 & 51.26 & 70.04 & 65.92 & +5.24 [-4.10, +15.09]\\
LLaVA & 40.32 [36.08, 44.66] & 67.52 & 55.36 & 77.78 & 78.63 & +11.11 [+2.12, +20.13]\\
Qwen2.5-VL & 43.18 [38.67, 47.82] & 65.16 & 52.65 & 81.29 & 80.00 & +14.84 [+8.12, +22.08]\\
Qwen3-VL & 43.71 [39.65, 47.66] & 67.65 & 57.05 & 80.78 & 80.20 & +12.55 [+6.71, +18.20]\\
\bottomrule\end{tabular}
\end{table*}

The advantage of maximum-first does not depend on a particular array. Across the 64 designs, discovery at 25 queries ranges from 43.82\% to 59.55\% for BLIP-2, 47.01\% to 64.10\% for LLaVA, 40.65\% to 61.29\% for Qwen2.5-VL and 47.06\% to 67.06\% for Qwen3-VL, and maximum-first exceeds the best design for every model.

Historical orders matter most when budgets are tight (\figref{fig:transfer}). At 10 queries, other-model history already finds about 70\% of affected Qwen questions, whereas maximum-first finds about 30\%; the curves converge by 200 queries. Better discovery also yields cleaner acceptance: at 25 queries, residual failure among accepted questions falls from 12.30\% under maximum-first to 7.45\% under other-model history for Qwen2.5-VL, and from 10.13\% to 6.45\% for Qwen3-VL.

The transfer also holds across datasets. In leave-one-dataset-out replay with calibration from the other two sources, same-model frequency improved 25-query discovery in 11 of the 12 model--dataset comparisons and tied in the remaining one.

The natural setting for these orders is model iteration. When a model is updated or replaced, the grids of earlier models already exist, and the new model needs only screening plus 25 joint queries per root, about 10,000--13,000 calls for 300 roots. Acquiring a comparable history from scratch would cost 62,000--154,000 same-model or 327,000--420,000 other-model calls per split (Appendix~\ref{app:cost}), so the orders are best viewed as a way to amortize completed evaluations across successive model versions.

\section{Discussion}
The coverage audit turns an acceptance decision into a question about specific requests: which tests passed, which combinations were omitted, and which accepted requests fail in the rest of the library. Its central lesson is that event-level robustness and request-level robustness diverge. A model can fail fewer than one joint record in a hundred and still leave a quarter of its single-test-passing requests vulnerable, because failures concentrate on particular requests and particular off-corner coordinates. Reports that average over conditions cannot reveal this; aligned grids can. The failure evidence is also stable: re-execution reproduces about 90\% of the archived failures while fewer than 1\% of the matched passing controls flip, so the gap reflects the inputs rather than run-to-run variance.

This has direct consequences for test design. Maximum-strength corners are a reasonable first gate but not a sufficient one, since each strength level is a separate realization and nearby off-corner inputs expose different failures. Magnitude coverage and realization coverage are therefore separate design choices, and the grid makes both explicit. When budgets are limited, completed evaluations of earlier models are the most informative guide to where a new model will fail, which makes the audit particularly suited to iterative development: the replacement audit in Appendix~\ref{app:changes} follows the successes users already rely on through a model update.

For model development, the audit supplies concrete targets. Family-pair hotspots such as blur with keyboard substitutions identify combinations for regression suites and augmentation, and matched channel profiles identify which modality loses established success first.

\subsection{Limitations}
The reported fractions are defined relative to the evaluated library: 1,500 questions from three VQA sources, four open-weight models of 2.7--8B parameters, and five families with four strength levels per channel. Larger libraries, further realizations or larger models may change the absolute fractions. Success is scored by a fixed automated protocol, and the re-execution study repeats the archived failure coordinates and matched controls rather than the full grid. Scene-text datasets are not part of the library, so robustness to degradation of text inside the image is outside this study, and the semantic-preservation audit samples 300 perturbed questions rather than every joint input.

\section{Conclusion}
M$^3$-Test links multimodal robustness testing to the evidence required for acceptance. Fully aligned response grids show that, for the Qwen models, over a quarter of the requests passing all single-modality tests fail jointly, and that 11--14\% of the requests passing maximum-strength corners still fail elsewhere. The same grids make completed evaluations reusable: risk orders learned from other models discover about 80\% of affected requests within 25 queries. After a test suite passes, M$^3$-Test answers which successful requests remain vulnerable and how further testing can expose them.

\section*{Data and Code Availability}
The response grids, analysis code and evaluation protocols needed to reproduce the reported tables and figures, together with the answer-review records and the re-execution evidence, are provided in Supplementary File 1. Analysis code and released result tables are at \url{https://github.com/Tender221/M3-bench}.
\section*{Declaration of Competing Interest}
The authors declare no competing interests.

\FloatBarrier
\appendix
\section{Input Construction and Calibration}\label{app:operators}
Table~\ref{tab:perturbation_implementation} specifies the unchanged input library. Each level has one deterministic realization. Protected text tokens retain task anchors; probability-based edits can leave short questions unchanged. The asset seed incorporates the original image, family and level, so brightness and contrast directions can differ across levels. These choices make actual input identity relevant to cost accounting. The semantic-preservation audit draws five questions from each text family, level and source dataset, giving 300 items on which two annotators independently judged meaning preservation, answer intent and readability; they agreed on every item and the released summary reports Wilson 95\% intervals for each stratum.
\begin{table*}[!t]
\centering
\caption{Frozen perturbation implementation contract.}
\label{tab:perturbation_implementation}
\fontsize{9}{10.8}\selectfont
\setlength{\tabcolsep}{3.5pt}
\renewcommand{\arraystretch}{1.12}
\begin{tabularx}{\textwidth}{@{}>{\raggedright\arraybackslash}p{0.12\textwidth}>{\raggedright\arraybackslash}p{0.16\textwidth}>{\raggedright\arraybackslash}p{0.27\textwidth}>{\raggedright\arraybackslash}X@{}}
\toprule
\textbf{Family} & \textbf{Operator} & \textbf{$S_1/S_2/S_3/S_4$ control} & \textbf{Deterministic realization} \\
\midrule
Gaussian blur & Gaussian blur & radius $1/2/4/6$ & Pillow GaussianBlur; clean images are returned unchanged at $S_0$. \\
Gaussian noise & Additive normal noise & $\sigma=5/10/20/30$ intensity units & NumPy seeded generator; add in the $[0,255]$ domain, clip, and cast to uint8. \\
JPEG & Lossy re-encoding & quality $70/40/20/10$ & Pillow JPEG encoding; the perturbed asset remains JPEG at the selected quality. \\
Brightness & Global luminance factor & $\delta=.15/.30/.45/.60$ & Pillow brightness factor $1\pm\delta$; sign is selected deterministically from the asset seed. \\
Contrast & Global contrast factor & $\delta=.15/.30/.45/.60$ & Pillow contrast factor $1\pm\delta$; sign uses a decorrelated asset seed. \\
Typo & Token-level edit & $p=.08/.18/.32/.48$ & One internal character is swapped, duplicated, dropped, or case-toggled; protected tokens are skipped. \\
Keyboard & Keyboard-neighbor substitution & $p=.08/.18/.32/.48$ & One internal character is replaced by a QWERTY neighbor; protected tokens are skipped. \\
OCR-like & Token/character confusion & $(p_{\rm tok},p_{\rm char})=$\newline $(.12,.30)/(.22,.40)/$\newline $(.34,.50)/(.48,.60)$ & Select eligible tokens with $p_{\rm tok}$; within each selected token, replace eligible characters with $p_{\rm char}$. Confusions include $o\!\to\!0$, $l/i\!\to\!1$, $s\!\to\!5$ and $e\!\to\!3$. \\
Whitespace/\allowbreak punctuation & Surface formatting noise & space jitter $=.06/.14/.24/.34$; punctuation drop $=0/.12/.25/.45$; comma insertion $=0/.02/.05/.08$ & Single spaces may double; punctuation is dropped or light commas inserted with the listed probabilities. \\
Casing & Character case toggle & $p=.01/.03/.06/.10$ & Each alphabetic character is toggled independently with the listed probability. \\
\bottomrule
\end{tabularx}
\end{table*}

\begin{figure*}[pos=tbp]
\centering\includegraphics[width=.92\textwidth]{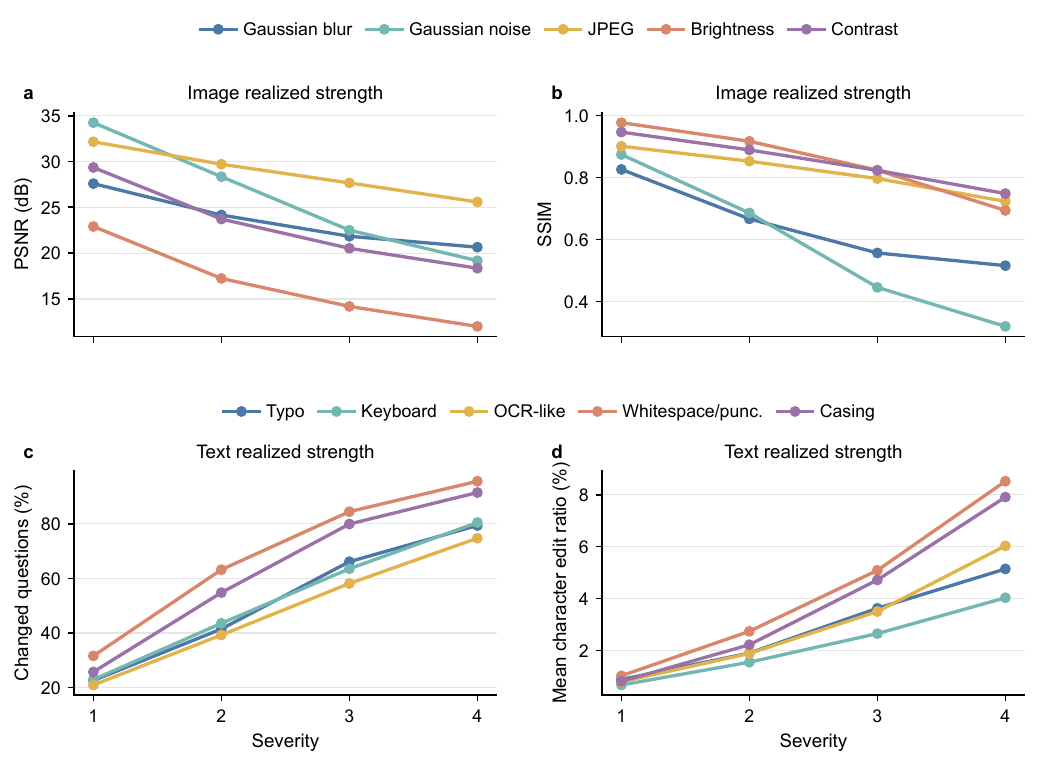}
\caption{Input-magnitude calibration, independent of answer grading. PSNR/SSIM summarize 21,300 original-image comparisons; changed-question rate and character-edit ratio summarize 30,000 text variants. Family-wise trajectories order realized magnitude on average.}
\label{fig:calibration}
\end{figure*}

\section{Scoring and Actual-Input Controls}\label{app:controls}
Semantic sensitivity uses both semantic requalification and semantic outcomes on fixed Combined eligibility, separating grading effects from population changes. The Judge is a Qwen3-4B-Instruct-2507 model, so it comes from the same model family as one evaluated system; every model is scored with the same reference-based questions, and Section~\ref{app:repeat} quantifies the residual grading error on the most fragile failure records.

The affected-request and residual-failure patterns remain under semantic-only grading (Table~\ref{tab:semantic_coverage}). Each row uses its own stated qualification and accepted population. Fixed-cohort semantic scoring preserves the questions passing the combined criterion while applying semantic outcomes to their corners and remaining joint conditions.
\begin{table*}[pos=tbp]
\centering\small\setlength{\tabcolsep}{5pt}
\caption{Coverage sensitivity to answer grading for the two Qwen models. Affected divides questions failing anywhere by the qualified cohort; residual divides failures outside the 25 nominal corners by the questions passing those corners. Combined requires semantic correctness and format compliance. Semantic requalification applies semantic-only success to qualification and outcomes; fixed-cohort semantic retains the original Combined qualification.}
\label{tab:semantic_coverage}
\begin{tabular}{@{}llrrr@{}}\toprule
Model & Scoring and screening & Passed($n$) & Affected(\%) & Residual after corners(\%)\\\midrule
Qwen2.5-VL & Combined & 540 & 28.70 & 14.06\\
Qwen2.5-VL & Semantic requalified & 571 & 29.77 & 15.40\\
Qwen2.5-VL & Semantic fixed-cohort & 540 & 28.15 & 13.97\\
Qwen3-VL & Combined & 658 & 25.84 & 11.43\\
Qwen3-VL & Semantic requalified & 659 & 25.80 & 11.41\\
Qwen3-VL & Semantic fixed-cohort & 658 & 25.84 & 11.43\\\bottomrule
\end{tabular}
\end{table*}

Equivalent input groups are defined within each question by RGB dimensions/pixels and the exact question string; there are 37,842 multi-entry groups per model. Responses within a group differ in 644/117/2,997/655 groups for BLIP-2/LLaVA/Qwen2.5/Qwen3, and success labels conflict in 154/51/619/227. Identical answers always receive identical semantic, format and combined labels.

The changed-channel, label-consistent control in Table~\ref{tab:input_control} retains the original qualification. All four affected-question sets match their nominal counterparts, as do the maximum-corner and diagonal omission sets; collapsing equivalent joint inputs changes only event denominators. Distinct-input replay charges each unknown group once and treats groups already queried during qualification as known.
\begin{table*}[pos=tbp]
\centering\small\setlength{\tabcolsep}{5pt}
\caption{Both-channels-changed, success-consistent sensitivity on the fixed cohorts passing clean and all singles. Eligible and failed columns count retained nominal records; the final columns count distinct questions and original images.}
\label{tab:input_control}
\begin{tabular}{@{}lrrrrr@{}}\toprule
Model & Passed($n$) & Eligible records & Failed records & Affected questions & Affected images\\\midrule
BLIP-2 & 145 & 29636 & 605 & 89 & 81\\
LLaVA & 550 & 121180 & 660 & 117 & 114\\
Qwen2.5-VL & 540 & 119610 & 1066 & 155 & 153\\
Qwen3-VL & 658 & 147300 & 1089 & 170 & 162\\
\bottomrule\end{tabular}
\end{table*}

\subsection{Answer review and execution repeatability}\label{app:repeat}
\begin{table*}[pos=htbp]
\centering
\small
\setlength{\tabcolsep}{5pt}
\caption{Execution repeatability of archived joint failures. Every archived joint failure of each model was re-executed once on the original environment, together with an equal number of matched passing controls drawn from the same questions. The last column counts the questions missed by the maximum-strength corners that failed again, with parentheses giving the subset whose three anchors stayed correct.}
\label{tab:repeat}
\begin{tabular}{@{}lrrrrr@{}}
\toprule
Model & \shortstack{Failures\\reproduced(\%)} & \shortstack{Matched controls\\flipped(\%)} & \shortstack{Affected questions\\still failing} & \shortstack{Failures with\\anchors correct(\%)} & \shortstack{Corner-missed\\rediscovered} \\
\midrule
Qwen2.5-VL & 89.83 (998/1,111) & 0.27 (3/1,111) & 97.42 (151/155) & 84.43 (938/1,111) & 59/63 (59) \\
Qwen3-VL   & 90.44 (993/1,098) & 0.64 (7/1,098) & 94.71 (161/170) & 85.61 (940/1,098) & 55/63 (52) \\
\bottomrule
\end{tabular}
\end{table*}

The answer review targets the questions whose affected status rests on a single failed joint coordinate, where one judging error changes the outcome: 37 such questions for Qwen2.5-VL and 36 for Qwen3-VL, giving 71 records after removing two entries that repeat one packing question across the two models, together with 20 matched passing controls (10 per model). Each record was re-examined against the image, question, reference answers and archived response, and the decision was recorded on the existing structured audit record. Retaining the original labels leaves 63 failures (88.7\%). Resolving synonyms, reference-supported alternatives and answer-format disagreements in favor of the recorded evidence leaves 57 records supported as valid joint failures (80.3\%) and 6 consistency disputes (8.5\%); the remaining 8 records comprise 4 answers that the Judge should have accepted and 4 input or eligibility issues. Of the 20 controls, 16 pass as expected, 2 raise the same input or eligibility issue and 2 remain undecided. The cohort therefore audits the most fragile failure evidence rather than a random sample, so it measures the precision of failure labels on these records rather than the overall accuracy of the Judge.

The repeatability study re-executed every archived joint failure of the two Qwen models together with an equal number of matched passing controls drawn from the same questions. Runs kept the original environment, model configurations, processors, prompt, greedy decoding and 32 new-token limit, with one request per batch instead of the original 6 or 4. This required 9,342 vision-model calls; all 9,342 original and 9,342 new answers were then graded in one pass under the same frozen binary protocol, and the re-graded original labels agreed exactly with the archived ones. Table~\ref{tab:repeat} reports the outcome.

Clean answers passed again for every affected question (155/155 and 170/170), and the distinct single anchors passed 2,170/2,176 and 2,415/2,423. These anchor changes are retained and do not alter the qualification set or the denominators. Reproduction is a targeted check of the archived failure coordinates rather than a re-run of the full grid, so the coverage fractions reported in the main text remain those of the original library.

\subsection{Accepted and residual populations}\label{app:gate_counts}
Table~\ref{tab:gate_counts} gives the counts underlying the residual fractions. For cohort size $n_Q$, affected questions $n_A$ and questions discovered by the gate $n_D$, the accepted denominator is $n_Q-n_D$ and the residual numerator is $n_A-n_D$; the original-image bootstrap resamples both together.
\begin{table*}[pos=tbp]
\centering\small\setlength{\tabcolsep}{5pt}
\caption{Finite-library residual failure among passes. The gates add 25 or 100 nominal joint conditions after clean and all 40 singles.}
\label{tab:gate_counts}
\begin{tabular}{@{}lrrrrrr@{}}\toprule
 & \multicolumn{3}{c}{Maximum corners} & \multicolumn{3}{c}{Diagonals}\\
Model & Residual($n$) & Accepted($n$) & Risk(\%) & Residual($n$) & Accepted($n$) & Risk(\%)\\\midrule
BLIP-2 & 41 & 97 & 42.27 & 24 & 80 & 30.00\\
LLaVA & 42 & 475 & 8.84 & 19 & 452 & 4.20\\
Qwen2.5-VL & 63 & 448 & 14.06 & 34 & 419 & 8.11\\
Qwen3-VL & 63 & 551 & 11.43 & 37 & 525 & 7.05\\
\bottomrule\end{tabular}
\end{table*}

\section{Profile Sensitivity and Full-Cost Accounting}\label{app:profiles}
Table~\ref{tab:profiles} uses 469 clean questions and 141,396 eligible joint records shared by all four models under the combined correctness-and-form criterion. Semantic requalification yields 516 questions and 158,283 joints; the fixed-cohort view retains the original supports.
\begin{table*}[pos=tbp]
\centering\small\setlength{\tabcolsep}{4pt}
\caption{Matched retention scores, percentage estimate [original-image 95\% interval]. Fixed-cohort semantic changes outcome grading without changing eligibility.}
\label{tab:profiles}
\begin{tabular}{@{}llrrr@{}}\toprule
Model & View & Image retention & Text retention & Joint retention\\\midrule
BLIP-2 & Combined & 90.31 [88.71, 91.82] & 88.61 [87.32, 89.79] & 97.75 [97.47, 98.03]\\
LLaVA & Combined & 98.61 [98.19, 99.01] & 97.48 [96.88, 98.00] & 99.78 [99.70, 99.86]\\
Qwen2.5-VL & Combined & 96.10 [95.09, 96.96] & 97.97 [97.43, 98.46] & 99.38 [99.18, 99.54]\\
Qwen3-VL & Combined & 98.28 [97.72, 98.76] & 98.34 [97.88, 98.75] & 99.75 [99.66, 99.84]\\
BLIP-2 & Semantic & 90.79 [89.34, 92.20] & 90.02 [88.87, 91.06] & 97.84 [97.57, 98.10]\\
LLaVA & Semantic & 98.33 [97.74, 98.80] & 97.32 [96.73, 97.85] & 99.72 [99.61, 99.81]\\
Qwen2.5-VL & Semantic & 96.08 [95.10, 96.92] & 98.06 [97.57, 98.51] & 99.42 [99.24, 99.57]\\
Qwen3-VL & Semantic & 98.33 [97.81, 98.76] & 98.46 [98.03, 98.83] & 99.76 [99.67, 99.84]\\
BLIP-2 & Fixed-cohort semantic & 91.52 [89.99, 92.91] & 90.46 [89.31, 91.53] & 98.11 [97.85, 98.36]\\
LLaVA & Fixed-cohort semantic & 98.61 [98.19, 99.01] & 97.48 [96.88, 98.00] & 99.79 [99.71, 99.86]\\
Qwen2.5-VL & Fixed-cohort semantic & 96.25 [95.26, 97.11] & 98.28 [97.78, 98.73] & 99.45 [99.27, 99.60]\\
Qwen3-VL & Fixed-cohort semantic & 98.28 [97.72, 98.76] & 98.35 [97.90, 98.76] & 99.75 [99.66, 99.84]\\
\bottomrule\end{tabular}
\end{table*}

For LLaVA and Qwen2.5, the weak-channel directions persist under semantic requalification. Qwen3's common-cohort channel difference spans zero in all views. BLIP-2's Combined text-minus-image difference is positive, while its semantic-only interval spans zero; its stronger Combined direction therefore includes response-form sensitivity. All six pairwise model contrasts, own-qualified profiles and family-pair scores accompany the source tables.

\subsection{Sensitivity to the inspected family library}\label{app:library_size}
\begin{figure*}[pos=tbp]
\centering\includegraphics[width=\textwidth]{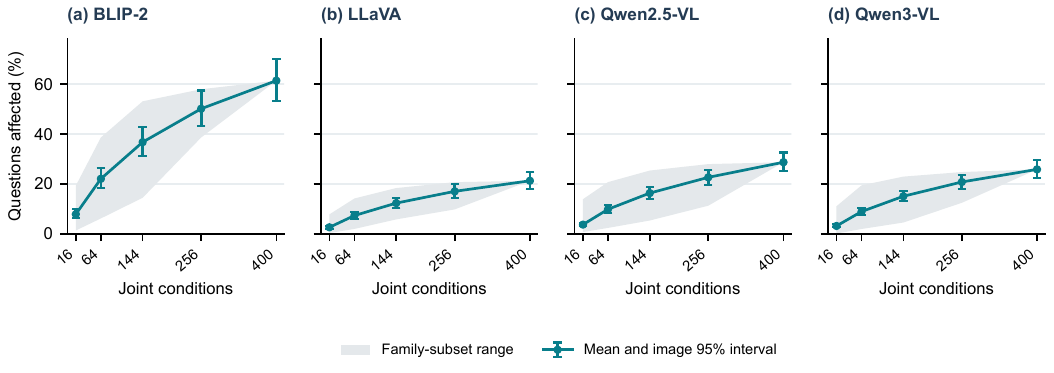}
\caption{Affected fraction as the inspected family library expands. Balanced sublibraries include $r$ image and $r$ text families with all strength combinations, giving $16r^2$ joint conditions for $r=1,\ldots,5$. All family choices are enumerated at each size. Teal points are mean affected fractions on the fixed original cohort passing clean and all singles, with original-image bootstrap 95\% intervals. Gray bands show the minimum--maximum across family choices, a composition range rather than a confidence interval. The full-library endpoints match Table~\ref{tab:summary}.}
\label{fig:library_size}
\end{figure*}

Figure~\ref{fig:library_size} retains the full cohort passing clean and all singles while varying the families represented in the joint library. Balanced three-family subsets expose mean affected fractions of 16.35\% for Qwen2.5-VL and 15.08\% for Qwen3-VL, compared with 28.70\% and 25.84\% for the full library. The family-composition ranges show why the identity of tested operators matters at a fixed library size. Image-bootstrap intervals describe a separate source of variation: which original-image groups populate the benchmark. All 961 sublibraries are enumerated; exported summaries report each family-count combination, its composition range and changed-input sensitivity.

\subsection{Calibration and screening costs}\label{app:cost}
Table~\ref{tab:cost} averages the 15 fitted-fold scenarios: three seeds and five folds, each with about 1,200 calibration and 300 deployment roots. Screening includes rejected roots. Source-model calibration is charged even when target-model calibration joints are absent.
\begin{table}[pos=htbp]
\centering\footnotesize\setlength{\tabcolsep}{3pt}
\caption{Mean distinct-input VLM queries per fold at budget 25, rounded. Add calibration to deployment when acquiring history anew. The accounting unit is a distinct input query to a tested VLM; acquisition and deployment columns report this unit separately.}
\label{tab:cost}
\begin{tabular}{@{}lrrr@{}}\toprule
Target & \shortstack{Same-model\\calibration} & \shortstack{Other-three\\calibration} & \shortstack{Deployment\\screen + joints}\\\midrule
BLIP-2 & 61,751 & 419,542 & 10,383\\
LLaVA & 133,511 & 347,782 & 12,408\\
Qwen2.5-VL & 132,183 & 349,110 & 12,358\\
Qwen3-VL & 153,847 & 327,446 & 12,948\\
\bottomrule\end{tabular}
\end{table}

The source tables retain all budgets, single-source transfers, nominal costs and greedy orders. Intervals use the fixed-order image resampling defined in Section~\ref{sec:uncertainty}.

\section{Model Replacement as an Acceptance Audit}\label{app:changes}
Reference-success accounting connects joint testing to model replacement. Among the 385 questions for which Qwen2.5-VL passed all 441 entries, Qwen3-VL retained whole-library success on 307. The remaining 20.26\% first failed at clean, single or joint stages, with 10, 32 and 36 questions, respectively. Following the negative-flip perspective~\cite{Yan2021PositiveCongruent}, the fixed reference cohort keeps every lost success visible: a newer, more accurate model still breaks one in five of the requests its predecessor handled under every condition.

\renewcommand{\bibfont}{\fontsize{8.4pt}{9.3pt}\selectfont}
\setlength{\bibsep}{1pt}
\bibliographystyle{cas-model2-names}
\bibliography{references}

@inproceedings{shah2019cycle,
  title     = {Cycle-Consistency for Robust Visual Question Answering},
  author    = {Shah, Meet and Chen, Xinlei and Rohrbach, Marcus and Parikh, Devi},
  booktitle = {Proceedings of the IEEE/CVF Conference on Computer Vision and Pattern Recognition (CVPR)},
  pages     = {6649--6658},
  year      = {2019},
  doi       = {10.1109/CVPR.2019.00681}
}

@inproceedings{ishmam2025visual,
  title     = {Visual Robustness Benchmark for Visual Question Answering (VQA)},
  author    = {Ishmam, Md Farhan and Tashdeed, Ishmam and Saadat, Talukder Asir and Ashmafee, Md Hamjajul and Kamal, Abu Raihan Mostofa and Hossain, Md Azam},
  booktitle = {Proceedings of the IEEE/CVF Winter Conference on Applications of Computer Vision (WACV)},
  pages     = {6623--6633},
  year      = {2025},
  doi       = {10.1109/WACV61041.2025.00645}
}

@article{ma2024robust,
  title   = {Robust Visual Question Answering: Datasets, Methods, and Future Challenges},
  author  = {Ma, Jie and Wang, Pinghui and Kong, Dechen and Wang, Zewei and Liu, Jun and Pei, Hongbin and Zhao, Junzhou},
  journal = {IEEE Transactions on Pattern Analysis and Machine Intelligence},
  volume  = {46},
  pages   = {5575--5594},
  year    = {2024},
  doi     = {10.1109/TPAMI.2024.3366154}
}

@inproceedings{Guo2017Calibration,
  author    = {Chuan Guo and Geoff Pleiss and Yu Sun and Kilian Q. Weinberger},
  title     = {On Calibration of Modern Neural Networks},
  booktitle = {Proceedings of the 34th International Conference on Machine Learning},
  year      = {2017},
  pages     = {1321--1330}
}

@inproceedings{Dancette2023SelectiveVQA,
  author    = {Dancette, Corentin and Whitehead, Spencer and Maheshwary, Rishabh and Vedantam, Ramakrishna and Scherer, Stefan and Chen, Xinlei and Cord, Matthieu and Rohrbach, Marcus},
  title     = {Improving Selective Visual Question Answering by Learning From Your Peers},
  booktitle = {Proceedings of the IEEE/CVF Conference on Computer Vision and Pattern Recognition (CVPR)},
  pages     = {24049--24059},
  year      = {2023}
}

@inproceedings{Khan2024SelectiveVQA,
  author    = {Khan, Zaid and Fu, Yun},
  title     = {Consistency and Uncertainty: Identifying Unreliable Responses From Black-Box Vision-Language Models for Selective Visual Question Answering},
  booktitle = {Proceedings of the IEEE/CVF Conference on Computer Vision and Pattern Recognition (CVPR)},
  pages     = {10854--10863},
  year      = {2024}
}

@article{Sterz2025DARE,
  author  = {Sterz, Hannah and Pfeiffer, Jonas and Vuli{\'c}, Ivan},
  title   = {{DARE}: Diverse Visual Question Answering with Robustness Evaluation},
  journal = {Transactions of the Association for Computational Linguistics},
  volume  = {13},
  pages   = {1121--1145},
  year    = {2025},
  doi     = {10.1162/tacl.a.29}
}

@inproceedings{Chen2023AdaRobustness,
  author    = {Chen, Shuo and Gu, Jindong and Han, Zhen and Ma, Yunpu and Torr, Philip H. S. and Tresp, Volker},
  title     = {Benchmarking Robustness of Adaptation Methods on Pre-trained Vision-Language Models},
  booktitle = {Advances in Neural Information Processing Systems (NeurIPS)},
  volume    = {36},
  year      = {2023},
  doi       = {10.52202/075280-2255}
}

@inproceedings{Bhagwatkar2024VLMRobustness,
  title     = {Improving Adversarial Robustness in Vision-Language Models with Architecture and Prompt Design},
  author    = {Bhagwatkar, Rishika and Nayak, Shravan and Bashivan, Pouya and Rish, Irina},
  booktitle = {Findings of the Association for Computational Linguistics: EMNLP 2024},
  pages     = {17003--17020},
  year      = {2024},
  doi       = {10.18653/v1/2024.findings-emnlp.990},
  url       = {https://aclanthology.org/2024.findings-emnlp.990/}
}

@article{Peng2024PolarityVQA,
  title   = {Robust visual question answering via polarity enhancement and contrast},
  author  = {Peng, Dahe and Li, Zhixin},
  journal = {Neural Networks},
  volume  = {179},
  pages   = {106560},
  year    = {2024},
  doi     = {10.1016/j.neunet.2024.106560}
}

@article{Cao2024LanguagePriors,
  title   = {Overcoming language priors in visual question answering with cumulative learning strategy},
  author  = {Cao, Ruibin and Li, Zhixin},
  journal = {Neurocomputing},
  volume  = {608},
  pages   = {128419},
  year    = {2024},
  doi     = {10.1016/j.neucom.2024.128419}
}

@inproceedings{Marino2019OKVQA,
  author    = {Kenneth Marino and Mohammad Rastegari and Ali Farhadi and Roozbeh Mottaghi},
  title     = {{OK}-{VQA}: A Visual Question Answering Benchmark Requiring External Knowledge},
  booktitle = {Proceedings of the IEEE/CVF Conference on Computer Vision and Pattern Recognition (CVPR)},
  year      = {2019},
  pages     = {3195--3204}
}

@inproceedings{Hudson2019GQA,
  author    = {Drew A. Hudson and Christopher D. Manning},
  title     = {{GQA}: A New Dataset for Real-World Visual Reasoning and Compositional Question Answering},
  booktitle = {Proceedings of the IEEE/CVF Conference on Computer Vision and Pattern Recognition (CVPR)},
  year      = {2019}
}

@inproceedings{Goyal2017VQAv2,
  author    = {Yash Goyal and Tejas Khot and Douglas Summers-Stay and Dhruv Batra and Devi Parikh},
  title     = {Making the v in {VQA} Matter: Elevating the Role of Image Understanding in Visual Question Answering},
  booktitle = {Proceedings of the IEEE Conference on Computer Vision and Pattern Recognition (CVPR)},
  year      = {2017}
}

@inproceedings{Li2021AdversarialVQA,
  author    = {Linjie Li and Jie Lei and Zhe Gan and Jingjing Liu},
  title     = {Adversarial {VQA}: A New Benchmark for Evaluating the Robustness of {VQA} Models},
  booktitle = {Proceedings of the IEEE/CVF International Conference on Computer Vision (ICCV)},
  year      = {2021},
  pages     = {2042--2051}
}

@inproceedings{Chen2020CSS,
  author    = {Long Chen and Xin Yan and Jun Xiao and Hanwang Zhang and Shiliang Pu and Yueting Zhuang},
  title     = {Counterfactual Samples Synthesizing for Robust Visual Question Answering},
  booktitle = {Proceedings of the IEEE/CVF Conference on Computer Vision and Pattern Recognition (CVPR)},
  year      = {2020},
  pages     = {10800--10809}
}

@inproceedings{Kant2021ContrastClassify,
  author    = {Yash Kant and Abhinav Moudgil and Dhruv Batra and Devi Parikh and Harsh Agrawal},
  title     = {Contrast and Classify: Training Robust {VQA} Models},
  booktitle = {Proceedings of the IEEE/CVF International Conference on Computer Vision (ICCV)},
  year      = {2021},
  pages     = {1604--1613}
}

@article{Qiu2024MMRobustness,
  author  = {Jielin Qiu and Yi Zhu and Xingjian Shi and Florian Wenzel and Zhiqiang Tang and Ding Zhao and Bo Li and Mu Li},
  title   = {Benchmarking Robustness of Multimodal Image-Text Models under Distribution Shift},
  journal = {Data-centric Machine Learning Research},
  volume  = {1},
  number  = {2},
  pages   = {1--56},
  url     = {https://data.mlr.press/assets/pdf/v01-2.pdf},
  year    = {2024}
}

@inproceedings{Li2023BLIP2,
  author    = {Junnan Li and Dongxu Li and Silvio Savarese and Steven Hoi},
  title     = {{BLIP}-2: Bootstrapping Language-Image Pre-training with Frozen Image Encoders and Large Language Models},
  booktitle = {Proceedings of the 40th International Conference on Machine Learning (ICML)},
  series    = {Proceedings of Machine Learning Research},
  volume    = {202},
  pages     = {19730--19742},
  year      = {2023}
}

@article{Bai2025Qwen25VL,
  author  = {Shuai Bai and Keqin Chen and Xuejing Liu and Jialin Wang and Wenbin Ge and Sibo Song and Kai Dang and Peng Wang and Shijie Wang and Jun Tang and Humen Zhong and Yuanzhi Zhu and Mingkun Yang and Zhaohai Li and Jianqiang Wan and Pengfei Wang and Wei Ding and Zheren Fu and Yiheng Xu and Jiabo Ye and Xi Zhang and Tianbao Xie and Zesen Cheng and Hang Zhang and Zhibo Yang and Haiyang Xu and Junyang Lin},
  title   = {Qwen2.5-{VL} Technical Report},
  journal = {arXiv preprint arXiv:2502.13923},
  year    = {2025}
}

@inproceedings{Antol2015VQA,
  title     = {VQA: Visual Question Answering},
  author    = {Antol, Stanislaw and Agrawal, Aishwarya and Lu, Jiasen and Mitchell, Margaret and Batra, Dhruv and Zitnick, C. Lawrence and Parikh, Devi},
  booktitle = {Proceedings of the IEEE International Conference on Computer Vision (ICCV)},
  year      = {2015},
  pages     = {2425--2433}
}

@inproceedings{Hendrycks2019ImageNetC,
  title     = {Benchmarking Neural Network Robustness to Common Corruptions and Perturbations},
  author    = {Hendrycks, Dan and Dietterich, Thomas G.},
  booktitle = {International Conference on Learning Representations (ICLR)},
  year      = {2019}
}

@inproceedings{Belinkov2018SyntheticNoise,
  title     = {Synthetic and Natural Noise Both Break Neural Machine Translation},
  author    = {Belinkov, Yonatan and Bisk, Yonatan},
  booktitle = {International Conference on Learning Representations (ICLR)},
  year      = {2018}
}

@inproceedings{Ebrahimi2018HotFlip,
  title     = {{HotFlip}: White-Box Adversarial Examples for Text Classification},
  author    = {Ebrahimi, Javid and Rao, Anyi and Lowd, Daniel and Dou, Dejing},
  booktitle = {Proceedings of the 56th Annual Meeting of the Association for Computational Linguistics (ACL)},
  year      = {2018},
  pages     = {31--36}
}

@inproceedings{Wei2019EDA,
  title     = {{EDA}: Easy Data Augmentation Techniques for Boosting Performance on Text Classification Tasks},
  author    = {Wei, Jason and Zou, Kai},
  booktitle = {Proceedings of the 2019 Conference on Empirical Methods in Natural Language Processing and the 9th International Joint Conference on Natural Language Processing (EMNLP-IJCNLP)},
  year      = {2019},
  pages     = {6382--6388}
}

@inproceedings{Karimi2021AEDA,
  title     = {{AEDA}: An Easier Data Augmentation Technique for Text Classification},
  author    = {Karimi, Akbar and Rossi, Leonardo and Prati, Andrea},
  booktitle = {Findings of the Association for Computational Linguistics: EMNLP 2021},
  year      = {2021},
  pages     = {2748--2754}
}

@inproceedings{Singh2019TextVQA,
  title     = {Towards {VQA} Models That Can Read},
  author    = {Singh, Amanpreet and Natarajan, Vivek and Shah, Meet and Jiang, Yu and Chen, Xinlei and Batra, Dhruv and Parikh, Devi and Rohrbach, Marcus},
  booktitle = {Proceedings of the IEEE/CVF Conference on Computer Vision and Pattern Recognition (CVPR)},
  year      = {2019},
  pages     = {8317--8326}
}

@inproceedings{Biten2019STVQA,
  author    = {Biten, Ali Furkan and Tito, Ruben and Mafla, Andres and Gomez, Lluis and Rusinol, Marcal and Valveny, Ernest and Jawahar, C. V. and Karatzas, Dimosthenis},
  title     = {Scene Text Visual Question Answering},
  booktitle = {Proceedings of the IEEE/CVF International Conference on Computer Vision (ICCV)},
  pages     = {4291--4301},
  year      = {2019},
  doi       = {10.1109/ICCV.2019.00439}
}

@article{Zhou2023SparseTextVQA,
  author  = {Zhou, Sheng and Guo, Dan and Li, Jia and Yang, Xun and Wang, Meng},
  title   = {Exploring Sparse Spatial Relation in Graph Inference for Text-Based {VQA}},
  journal = {IEEE Transactions on Image Processing},
  volume  = {32},
  pages   = {5060--5074},
  year    = {2023},
  doi     = {10.1109/TIP.2023.3310332}
}

@article{Zhou2024GPIN,
  author  = {Zhou, Sheng and Guo, Dan and Yang, Xun and Dong, Jianfeng and Wang, Meng},
  title   = {Graph Pooling Inference Network for Text-Based {VQA}},
  journal = {ACM Transactions on Multimedia Computing, Communications, and Applications},
  volume  = {20},
  number  = {4},
  pages   = {1--21},
  year    = {2024},
  doi     = {10.1145/3634918}
}

@inproceedings{Zhou2025EgoTextVQA,
  author    = {Zhou, Sheng and Xiao, Junbin and Li, Qingyun and Li, Yicong and Yang, Xun and Guo, Dan and Wang, Meng and Chua, Tat-Seng and Yao, Angela},
  title     = {{EgoTextVQA}: Towards Egocentric Scene-Text Aware Video Question Answering},
  booktitle = {Proceedings of the IEEE/CVF Conference on Computer Vision and Pattern Recognition (CVPR)},
  pages     = {3363--3373},
  year      = {2025}
}

@article{Zhou2026GroundedTextVQA,
  author  = {Zhou, Sheng and Xiao, Junbin and Yang, Xun and Song, Peipei and Guo, Dan and Yao, Angela and Wang, Meng and Chua, Tat-Seng},
  title   = {Scene-Text Grounding for Text-Based Video Question Answering},
  journal = {IEEE Transactions on Multimedia},
  volume  = {28},
  pages   = {1417--1430},
  year    = {2026},
  doi     = {10.1109/TMM.2025.3639956}
}

@inproceedings{Gurari2018VizWiz,
  title     = {VizWiz Grand Challenge: Answering Visual Questions From Blind People},
  author    = {Gurari, Danna and Li, Qing and Stangl, Abigale J. and Guo, Anhong and Lin, Chi and Grauman, Kristen and Luo, Jiebo and Bigham, Jeffrey P.},
  booktitle = {Proceedings of the IEEE/CVF Conference on Computer Vision and Pattern Recognition (CVPR)},
  year      = {2018},
  pages     = {3608--3617}
}

@inproceedings{Latif2026VLSRB,
  author    = {Latif, Sameer Shafayet and Shiper, Sadab and Kiran, K. M. Rahiduzzaman and Ishmam, Md Farhan and Hossain, Md Azam and Kamal, Abu Raihan Mostofa and Ashmafee, Md Hamjajul},
  title     = {Enhancing Vision Language Corruption Robustness using Cross-Distribution \& Prompted Denoisers},
  booktitle = {Proceedings of the IEEE/CVF Winter Conference on Applications of Computer Vision (WACV)},
  pages     = {5754--5765},
  year      = {2026}
}

@article{Kang2026SCEval,
  author  = {Kang, Zhaolu and Wu, Meixin and Xue, Yu and He, Yingjie and Shi, Qiming and Wei, Lei and Wang, Yidi and Xuan, Richeng and Hu, Zhichao},
  title   = {Modality Fault Lines: Structural Corruptions Reveal Fragile Omni-Modal Reasoning},
  journal = {arXiv preprint arXiv:2608.29278},
  year    = {2026},
  doi     = {10.48550/arXiv.2608.29278},
  url     = {https://arxiv.org/abs/2608.29278v2}
}

@inproceedings{Liu2024LLaVA15,
  author    = {Liu, Haotian and Li, Chunyuan and Li, Yuheng and Lee, Yong Jae},
  title     = {Improved Baselines with Visual Instruction Tuning},
  booktitle = {Proceedings of the IEEE/CVF Conference on Computer Vision and Pattern Recognition (CVPR)},
  pages     = {26296--26306},
  year      = {2024}
}

@article{Bai2025Qwen3VL,
  author = {Bai, Shuai and Cai, Yuxuan and Chen, Ruizhe and Chen, Keqin and others},
  title = {{Qwen3-VL} Technical Report},
  journal = {arXiv preprint arXiv:2511.21631},
  year = {2025},
  doi = {10.48550/arXiv.2511.21631},
  url = {https://arxiv.org/abs/2511.21631}
}

@article{Rosenfeld2025VQAStability,
  author = {Rosenfeld, Amir and Glazer, Neta and Fetaya, Ethan},
  title = {Questioning the Stability of Visual Question Answering},
  journal = {arXiv preprint arXiv:2511.11206},
  year = {2025},
  doi = {10.48550/arXiv.2511.11206},
  url = {https://arxiv.org/abs/2511.11206}
}

@inproceedings{Yan2021PositiveCongruent,
  author = {Yan, Sijie and Xiong, Yuanjun and Kundu, Kaustav and Yang, Shuo and Deng, Siqi and Wang, Meng and Xia, Wei and Soatto, Stefano},
  title = {Positive-Congruent Training: Towards Regression-Free Model Updates},
  booktitle = {Proceedings of the IEEE/CVF Conference on Computer Vision and Pattern Recognition (CVPR)},
  pages = {14299--14308},
  year = {2021},
  doi = {10.1109/CVPR46437.2021.01407}
}

@inproceedings{Jimenez2022CARETS,
  title = {{CARETS}: A Consistency And Robustness Evaluative Test Suite for {VQA}},
  author = {Jimenez, Carlos E. and Russakovsky, Olga and Narasimhan, Karthik},
  booktitle = {Proceedings of the 60th Annual Meeting of the Association for Computational Linguistics (Volume 1: Long Papers)},
  year = {2022},
  pages = {6392--6405},
  publisher = {Association for Computational Linguistics},
  doi = {10.18653/v1/2022.acl-long.443},
  url = {https://aclanthology.org/2022.acl-long.443/}
}

@inproceedings{Chou2025MMR3,
  title = {{MM-R}$^3$: On (In-)Consistency of Vision-Language Models ({VLMs})},
  author = {Chou, Shih-Han and Chandhok, Shivam and Little, Jim and Sigal, Leonid},
  booktitle = {Findings of the Association for Computational Linguistics: ACL 2025},
  year = {2025},
  pages = {4762--4788},
  publisher = {Association for Computational Linguistics},
  doi = {10.18653/v1/2025.findings-acl.246},
  url = {https://aclanthology.org/2025.findings-acl.246/}
}

@article{Shirnin2024Robustness,
  title = {Analyzing the Robustness of Vision \& Language Models},
  author = {Shirnin, Alexander and Andreev, Nikita and Potapova, Sofia and Artemova, Ekaterina},
  journal = {IEEE/ACM Transactions on Audio, Speech, and Language Processing},
  year = {2024},
  volume = {32},
  pages = {2751--2763},
  doi = {10.1109/TASLP.2024.3399061},
  url = {https://publications.hse.ru/en/articles/943901261}
}

@article{Duan2026VLTest,
  author = {Duan, Bin and Lin, Jintao and Yang, Guowei},
  title = {{VLTest}: Finding Joint Image--Text Failures in Vision--Language Models},
  journal = {arXiv preprint arXiv:2610.30794},
  year = {2026},
  url = {https://arxiv.org/abs/2610.30794}
}

@article{Weissl2026MoST,
  author = {Wei{\ss}l, Oliver and Breitembach, Eduardo and Stocco, Andrea},
  title = {Search-Based Testing of Modality Sensitivity in Vision Language Models},
  journal = {arXiv preprint arXiv:2610.32360},
  year = {2026},
  url = {https://arxiv.org/abs/2610.32360}
}

@inproceedings{Feng2020DeepGini,
  author = {Feng, Yang and Shi, Qingkai and Gao, Xinyu and Wan, Jun and Fang, Chunrong and Chen, Zhenyu},
  title = {{DeepGini}: Prioritizing Massive Tests to Enhance the Robustness of Deep Neural Networks},
  booktitle = {Proceedings of the 29th ACM SIGSOFT International Symposium on Software Testing and Analysis},
  year = {2020},
  url = {https://arxiv.org/abs/1903.00661}
}

@article{Weiss2022Simple,
  author = {Weiss, Michael and Tonella, Paolo},
  title = {Simple Techniques Work Surprisingly Well for Neural Network Test Prioritization and Active Learning (Replicability Study)},
  journal = {arXiv preprint arXiv:2205.00664},
  year = {2022},
  url = {https://arxiv.org/abs/2205.00664}
}

@techreport{Kuhn2010Combinatorial,
  author = {Kuhn, D. Richard and Kacker, Raghu N. and Lei, Yu},
  title = {Practical Combinatorial Testing},
  institution = {National Institute of Standards and Technology},
  number = {Special Publication 800-142},
  year = {2010},
  doi = {10.6028/NIST.SP.800-142},
  url = {https://www.nist.gov/publications/practical-combinatorial-testing}
}

@inproceedings{Ma2019DeepCT,
  author = {Ma, Lei and Juefei-Xu, Felix and Xue, Minhui and Li, Bo and Li, Li and Liu, Yang and Zhao, Jianjun},
  title = {{DeepCT}: Tomographic Combinatorial Testing for Deep Learning Systems},
  booktitle = {2019 IEEE 26th International Conference on Software Analysis, Evolution and Reengineering (SANER)},
  pages = {614--618},
  year = {2019},
  doi = {10.1109/SANER.2019.8668044},
  url = {https://lilicoding.github.io/papers/ma2019deepct.pdf}
}

@inproceedings{Ma2018DeepGauge,
  author = {Ma, Lei and Juefei-Xu, Felix and Zhang, Fuyuan and Sun, Jiyuan and Xue, Minhui and Li, Bo and Chen, Chunyang and Su, Ting and Li, Li and Liu, Yang and Zhao, Jianjun and Wang, Yadong},
  title = {{DeepGauge}: Multi-Granularity Testing Criteria for Deep Learning Systems},
  booktitle = {Proceedings of the 33rd ACM/IEEE International Conference on Automated Software Engineering},
  year = {2018},
  doi = {10.1145/3238147.3238202},
  url = {https://arxiv.org/abs/1803.07519}
}

@inproceedings{HarelCanada2020Coverage,
  author = {Harel-Canada, Fabrice and Wang, Lingxiao and Gulzar, Muhammad Ali and Gu, Quanquan and Kim, Miryung},
  title = {Is Neuron Coverage a Meaningful Measure for Testing Deep Neural Networks?},
  booktitle = {Proceedings of the 28th ACM Joint Meeting on European Software Engineering Conference and Symposium on the Foundations of Software Engineering},
  pages = {851--862},
  year = {2020},
  doi = {10.1145/3368089.3409754},
  url = {https://people.cs.vt.edu/~gulzar/assets/pdf/Neuron-Coverage-FSE20.pdf}
}

@inproceedings{Ribeiro2020CheckList,
  author = {Ribeiro, Marco Tulio and Wu, Tongshuang and Guestrin, Carlos and Singh, Sameer},
  title = {Beyond Accuracy: Behavioral Testing of {NLP} Models with {CheckList}},
  booktitle = {Proceedings of the 58th Annual Meeting of the Association for Computational Linguistics},
  pages = {4902--4912},
  year = {2020},
  doi = {10.18653/v1/2020.acl-main.442},
  url = {https://aclanthology.org/2020.acl-main.442/}
}

@inproceedings{Zheng2023LLMJudge,
  author = {Zheng, Lianmin and others},
  title = {Judging {LLM}-as-a-Judge with {MT}-Bench and Chatbot Arena},
  booktitle = {Advances in Neural Information Processing Systems},
  volume = {36},
  year = {2023},
  url = {https://arxiv.org/abs/2306.05685}
}

@inproceedings{Panickssery2024SelfPreference,
  author = {Panickssery, Arjun and Bowman, Samuel R. and Feng, Shi},
  title = {{LLM} Evaluators Recognize and Favor Their Own Generations},
  booktitle = {Advances in Neural Information Processing Systems},
  volume = {37},
  year = {2024},
  url = {https://arxiv.org/abs/2404.13076}
}
\end{document}